\pdfoutput=1
\documentclass[11pt]{article}

\usepackage[final]{acl}

\usepackage{times}
\usepackage{latexsym}

\usepackage[T1]{fontenc}
\usepackage[utf8]{inputenc}

\usepackage{microtype}

\usepackage{inconsolata}

\usepackage{graphicx}
\usepackage{tabularx}
\usepackage{enumerate}
\usepackage{algorithm,algorithmic}
\usepackage{amsmath}
\usepackage{CJKutf8}
\usepackage{multirow}
\usepackage{arydshln} 
\usepackage[varbb]{newpxmath}
\DeclareUnicodeCharacter{FF5E}{~}
\usepackage{fix-cm}
\title{C-LLM: Learn to Check Chinese Spelling Errors Character by Character}

\author{Kunting Li\textsuperscript{1}\thanks{Work was done when Kunting Li was interning at WeChat AI, Tencent Inc, China.}, Yong Hu\textsuperscript{2}, Liang He\textsuperscript{1,3\thanks{Corresponding author.}}, Fandong Meng\textsuperscript{2}, Jie Zhou\textsuperscript{2} \\
  \textsuperscript{1}Department of Electronic Engineering, and Beijing National Research \\Center for Information Science and Technology, Tsinghua University, Beijing 100084, China\\
  \texttt{lkt22@mails.tsinghua.edu.cn, heliang@mail.tsinghua.edu.cn} \\
  \textsuperscript{2}WeChat AI, Tencent Inc, China \textsuperscript{3}Xinjiang University, Urumqi 830017, China \\
  \texttt{\{rightyonghu,fandongmeng,withtomzhou\}@tencent.com }
  \\
}

\begin{document}
\maketitle
\begin{abstract}
Chinese Spell Checking (CSC) aims to detect and correct spelling errors in sentences. Despite Large Language Models (LLMs) exhibit robust capabilities and are widely applied in various tasks, their performance on CSC is often unsatisfactory. We find that LLMs fail to meet the Chinese character-level constraints of the CSC task, namely equal length and phonetic similarity, leading to a performance bottleneck. Further analysis reveals that this issue stems from the granularity of tokenization, as current mixed character-word tokenization struggles to satisfy these character-level constraints. To address this issue, we propose C-LLM, a \textbf{L}arge \textbf{L}anguage \textbf{M}odel-based Chinese Spell \textbf{C}hecking method that learns to check errors \textbf{C}haracter by \textbf{C}haracter. Character-level tokenization enables the model to learn character-level alignment, effectively mitigating issues related to character-level constraints. Furthermore, CSC is simplified to replication-dominated and substitution-supplemented tasks. Experiments on two CSC benchmarks demonstrate that C-LLM achieves an average improvement of 10\% over existing methods. Specifically, it shows a 2.1\% improvement in general scenarios and a significant 12\% improvement in vertical domain scenarios, establishing state-of-the-art performance. The source code can be accessed at \url{https://github.com/ktlKTL/C-LLM}.
\end{abstract}

\section{Introduction}
\begin{figure}[!ht]
    \centering
    \includegraphics[width=\linewidth]{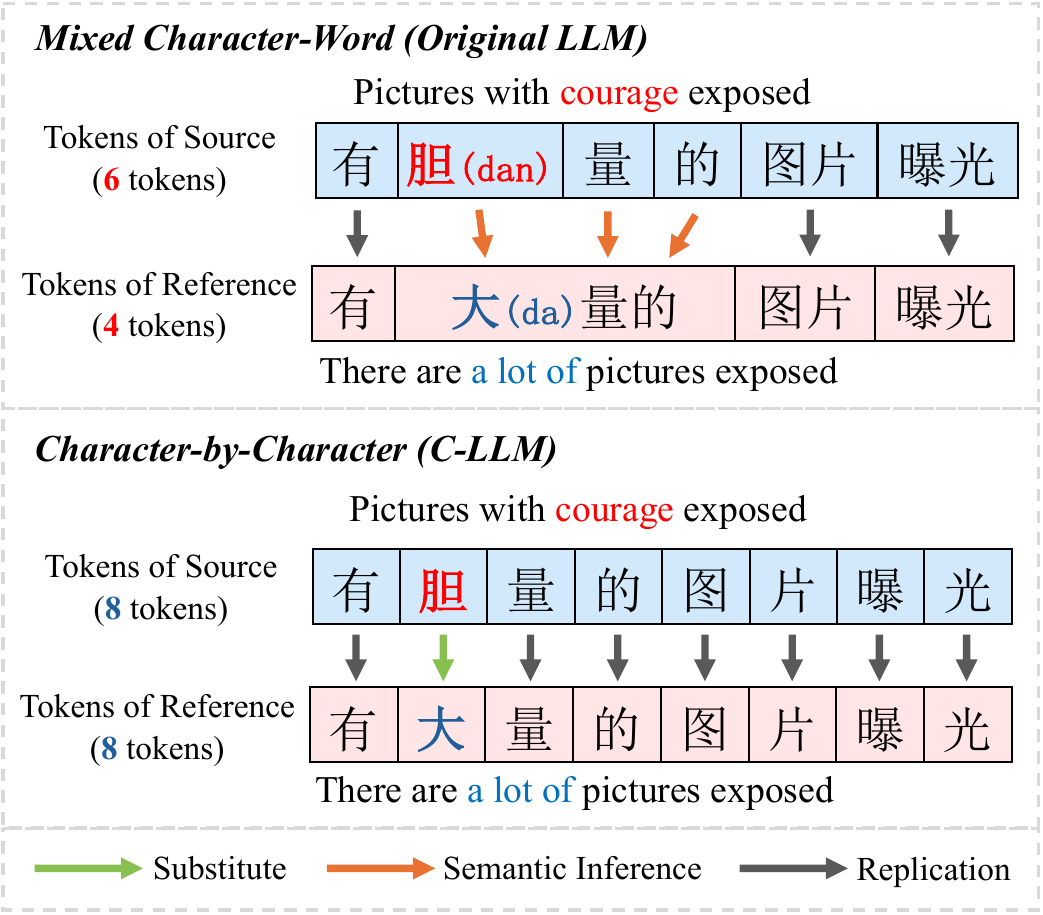}
    \caption{\label{fig:intro}Encoding differences between the original LLMs and C-LLM.} 
\end{figure}

Chinese Spell Checking (CSC) involves detecting and correcting erroneous characters in Chinese sentences, playing a vital role in applications \cite{gao2010large,yu2014chinese}. Although Large Language Models (LLMs) exhibit potent capabilities and are increasingly being applied to a variety of tasks \cite{wang2023chatgpt,he2023can,Grammar_chatgpt2}, previous studies \cite{li2021tail} showed that generative models, such as LLMs \cite{li2023effectiveness}, do not perform well on CSC.

The CSC task inherently involves character-level length and phonetic constraints. The character-level length constraint requires the predicted sentence maintain the same number of characters as the source sentence. Additionally, the phonetic constraint necessitates that the predicted characters closely match the phonetics of the source characters, as approximately 83\% of spelling errors are phonetically identical or similar to the correct ones \cite{liu-etal-2010-visually}. We find that LLMs often fail to meet these character-level length and phonetic constraints in the CSC task.

Using GPT-4 \cite{achiam2023gpt} as an example, we observed that under few-shot prompting, 10\% of the model's predicted sentences did not match the character count of the source sentences. In contrast, this issue was entirely absent in BERT-style models. Additionally, 35\% of predicted characters were phonetically dissimilar to the source characters, and errors due to non-homophone predictions account for approximately 70\% of all prediction errors. These deficiencies in character length and phonetic similarity result in outputs that fail to meet task requirements, leading to suboptimal correction performance.

\begin{CJK*}{UTF8}{gbsn} 

We find that the underlying issue lies in the granularity of the LLM's tokenization. The current mixed character-word tokenization results in a character-to-word mapping. This prevents LLMs from learning character-level alignment and tends to produce predictions that do not satisfy character-level constraints. As shown in Figure~\ref{fig:intro}, under the mixed character-word tokenization, the LLM needs to infer that multiple tokens corresponds to a single token (e.g., "\textit{胆(bold)}","\textit{大(large)}","\textit{的(of)}"->"\textit{大量的(large amount)}") and deduce implicit character alignment (e.g., "\textit{胆(bold)}"->"\textit{大(large)}"). These reasoning processes complicate the CSC, as the majority of CSC cases involve simply replicating characters. For example, the correct character "\textit{量(amount)}" is copied directly from the source. Despite the advancements in the semantic understanding capabilities of LLMs across various tasks, unclear character alignment can still lead to mis-corrections and over-corrections. Therefore, it is vital to establish explicit character-level alignment. 
\end{CJK*}

%These complicate the CSC， these 什么？ ✅

Building on this concept, we propose C-LLM, a \textbf{L}arge \textbf{L}anguage \textbf{M}odel-based Chinese Spell \textbf{C}hecking method that learns to check errors \textbf{C}haracter by \textbf{C}haracter. Our motivation is to encode at the character level and establish character-level alignment for training sentence pairs, thereby alleviating the issues related to character-level constraints. As shown in Figure～\ref{fig:intro}, this approach ensures that the number of tokens in sentence pairs remains consistent, making it easier for LLMs to learn the phonetic mappings between Chinese characters. Furthermore, CSC is simplified to the tasks of replicating correct characters and replacing incorrect ones, without complex reasoning.

% Under the premise that the source and reference sentences are of equal character-level length, training LLMs by mapping each character to a token can significantly simplify the task. 这句衔接逻辑不太好，不是说两个长度相等，所以构建字符映射能简化任务。 这句可以删去，或者移到上一段 These complicate the CSC, as the majority of CSC cases involve simply replicating characters.  附近。 ✅

Specifically, we construct the character-level tokenization to ensure that tokens are encoded according to individual Chinese characters. To adapt the model to the new vocabulary, we perform continued training on a large dataset. Furthermore, to enable the LLMs to learn CSC, we conduct supervised fine-tuning on CSC datasets. Experiments on the general dataset CSCD-NS \cite{hu2022CSCD} and the multi-domain dataset LEMON \cite{wu2023rethinking} show that C-LLM outperforms existing methods in both general and vertical domain scenarios, establishing state-of-the-art performance.

The contributions of this work can be summarized in three aspects: (1) We find that mixed character-word tokenization hinders LLM from effectively understanding the character-level constraints in CSC.
(2) We propose the C-LLM, which learns character-level alignment and can check errors character by character. (3) Through testing on general and multi-domain datasets, we found that C-LLM achieves state-of-the-art performance, providing insights for the design of future error correction models.
% analyze the performance of LLM in error correction and

\section{Related Work}
\label{sec:related work}
\textbf{BERT-style CSC Models} With the emergence of pre-trained language models, the dominant method for CSC has shifted to BERT-style models \cite{devlin-etal-2019-bert}, which treat CSC as a sequence labeling task. These models map each character in a sentence to its correct counterpart and are fine-tuned on pairs of sourece and reference sentences. Additionally, some studies have integrated phonological and morphological knowledge to improve the labeling process \cite{cheng-etal-2020-spellgcn,guo2021global,huang-etal-2021-phmospell,zhang2021correcting}. However, due to parameter constraints, these models underperform in low-frequency and complex semantic scenarios compared to LLMs.

\textbf{Autoregressive CSC models} 
Unlike BERT-style models, which can infer each token in parallel, autoregressive CSC models process tokens sequentially. Previous research \cite{li2021tail} indicates that autoregressive models like GPT-2 \cite{radford2019language} may underperform on CSC. With the advancement of LLMs, several studies have investigated their text correction capabilities. The study \cite{csc_chatgpt} finds that while ChatGPT \footnote{https://chat.openai.com} know the phonetics of Chinese characters, they can not understand how to pronounce it, making phonetic error correction challenging. Other studies \cite{Grammar_chatgpt1,Grammar_chatgpt2} note that ChatGPT often produces very fluent corrections but also introduces more over-corrections. These findings align with our observations, emphasizing the need to enhance LLMs' performance on CSC.

% underscoring 这个词比较奇怪 ✅

\section{Motivation}

\begin{table*}[!ht]
    \centering
    \resizebox{\linewidth}{!}{
    \begin{tabular}{l|ccc|ccc|ccc|ccc}
    \hline
        \multirow{3}{*}{Model} & \multicolumn{6}{c|}{Sentence Level (\%)} & \multicolumn{6}{c}{Character Level (\%)}  \\ \cline{2-13} 
        & \multicolumn{3}{c|}{Detection} & \multicolumn{3}{c|}{Correction} & \multicolumn{3}{c|}{Detection} & \multicolumn{3}{c}{Correction} \\ \cline{2-13} 
         & P & R & F1 & P & R & F1 & P & R & F1 & P & R & F1 \\ \hline
        GPT-4 & 58.50  & 60.23  & 59.35  & 53.35  & 54.93  & 54.13  & 58.52  & 65.78  & 61.94  & 51.41  & 57.79  & 54.41  \\ \hline
        BERT & 75.54  & 60.88  & 67.42  & 71.34  & 57.49  & 63.67  & 79.65  & 61.79  & 69.59  & 74.96  & 58.15  & 65.49  \\
        SMBERT & 75.68  & 62.96  & 68.74  & 71.45  & 59.44  & 64.90  & 79.97  & 64.12  & 71.17  & 75.53  & 60.56  & 67.22  \\
        SCOPE & \textbf{79.49}  & \textbf{66.96}  & \textbf{72.69}  & \textbf{76.39}  & \textbf{64.35}  & \textbf{69.86}  & \textbf{83.30}  & \textbf{68.08}  & \textbf{74.92}  & \textbf{79.72}  & \textbf{65.15}  & \textbf{71.70}  \\ \hline  
    \end{tabular}
    }
    \caption{\label{tab:gpt_cscd}The performance of GPT-4 and BERT-style models \cite{devlin-etal-2019-bert,zhang-etal-2020-spelling,li2022improving} on the CSCD-NS test set is evaluated at both the sentence and character levels, with precision (P), recall (R), and F1 score (F1) reported (\%) for both detection (D) and correction (C) tasks.}
\end{table*}
% ChatGPT & 63.29  & 49.94  & 55.83  & 58.34  & 46.03  & 51.46  & 64.08  & 53.64  & 58.40  & 57.61  & 48.22  & 52.50  \\

\subsection{Problem Formulation}
% 问题公式化
The CSC task aims to detect and correct all erroneous characters in Chinese sentence. Consider a source sentence $X_c = \left\{x_{c_1},x_{c_2},..,x_{c_n}\right\}$ consisting of $n$ characters, which may contain spelling errors. The corresponding reference sentence $Y_c = \left\{y_{c_1},y_{c_2},..,y_{c_n}\right\}$ contains the same number of characters as $X_c$, and with all errors corrected. Notably, a significant proportion of the corrected characters $y_{c_i}$ are phonetically identical or similar to erroneous character $x_{c_i}$. The CSC model identifies character-level spelling mistakes in the input $X_c$ and generates the predicted sentence $Y_c^{'} = \left\{y^{'}_{c_{1}},y^{'}_{c_{2}},..,y^{'}_{c_{m}}\right\}$, where $y^{'}_{c_i}$ is the character predicted for $x_{c_i}$ and $m$ should be equal to $n$ according to the CSC. In this process, the tokens of the source sentence and the reference sentence after tokenization can be represented as $X_t = \left\{x_{t_1},x_{t_2},...,x_{t_n} \right\}$ and $Y_t = \left\{y_{t_1},y_{t_2},...,y_{t_m} \right\}$, respectively.

% 这里X_{c}和X_{t}的写法要对应，X_{c} = \{ x_{c_{1}},...,x_{c_{n}}  \} ✅

\subsection{Analysis of LLMs in CSC}

LLMs now exhibit powerful language processing capabilities and are widely used \cite{zhao2023survey}. Similar to previous studies \cite{wang2023chatgpt,Grammar_chatgpt2}, we conduct a preliminary analysis of LLM performance on the CSC using GPT-4 \cite{achiam2023gpt} with in-context learning \cite{brown2020language}. Our experiments leverage the GPT-4 API and employ few-shot prompt (see Appendix~\ref{app:prompt_cscd}) on the CSCD-NS \cite{hu2022CSCD} test set for spelling correction. The prompt comprised five positive and five negative examples, randomly selected from the CSCD-NS training set.

As shown in Table~\ref{tab:gpt_cscd}, GPT-4's performance in spelling correction is inferior to that of BERT-style models. Our analysis indicates that GPT-4 struggles to meet two key constraints of the CSC task: character-level length and phonetic similarity. This misalignment results in a significant portion of the predictions that do not meet task requirements, leading to suboptimal correction performance.
\begin{figure}
    \centering
    \includegraphics[width=1\linewidth]{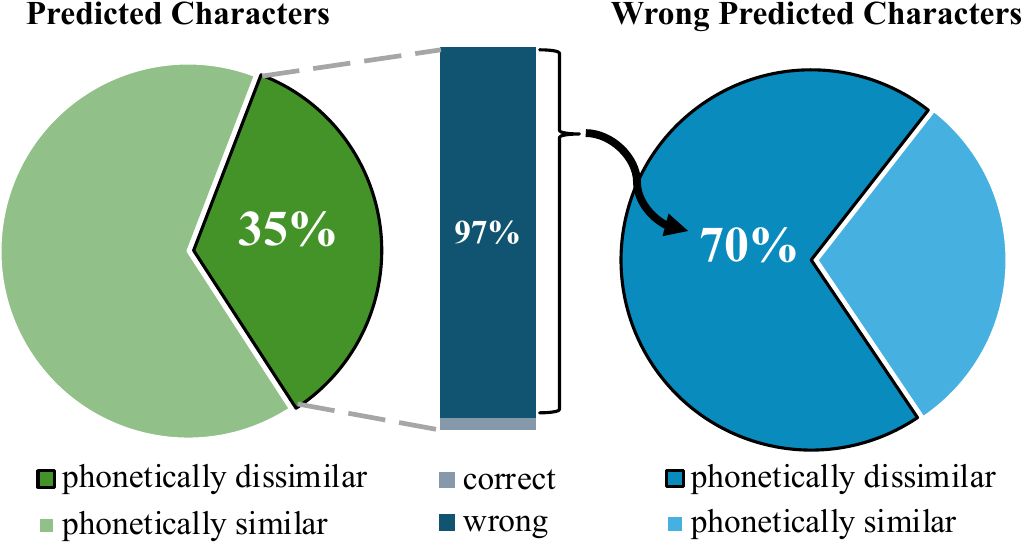}
    \caption{\label{fig:gpt_stat}Statistic results of non-homophone  characters.}
\end{figure}

% Specifically, the model frequently generates sentences with a different number of characters than the source and produces characters that are not phonetically similar to the source ones. 
Statistics reveal that 10\% of GPT-4's predicted sentences fail to meet the character-level length constraint, adversely affecting both precision and recall. Additionally, as illustrated in Figure~\ref{fig:gpt_stat}, GPT-4 generates 35\% of characters that are not phonetically similar to the source ones. Among these, 97\% are incorrect, and these incorrect phonologically dissimilar characters constitute a significant portion (70\%) of all prediction errors, severely impacting the model's performance. Therefore, identifying the root causes of LLMs' inability to satisfy character-level length and phonetic constraints is crucial for improving their performance.

% 这段跟intro有点重复，都是拿GPT4举例，数据也都一样。如果这里具体说了， intro可以再简单说（比如不提具体数值）。或者这里就要有更深度的分析数据。 ✅

\begin{figure*}[htbp]
    \centering
    \includegraphics[width=0.95\linewidth]{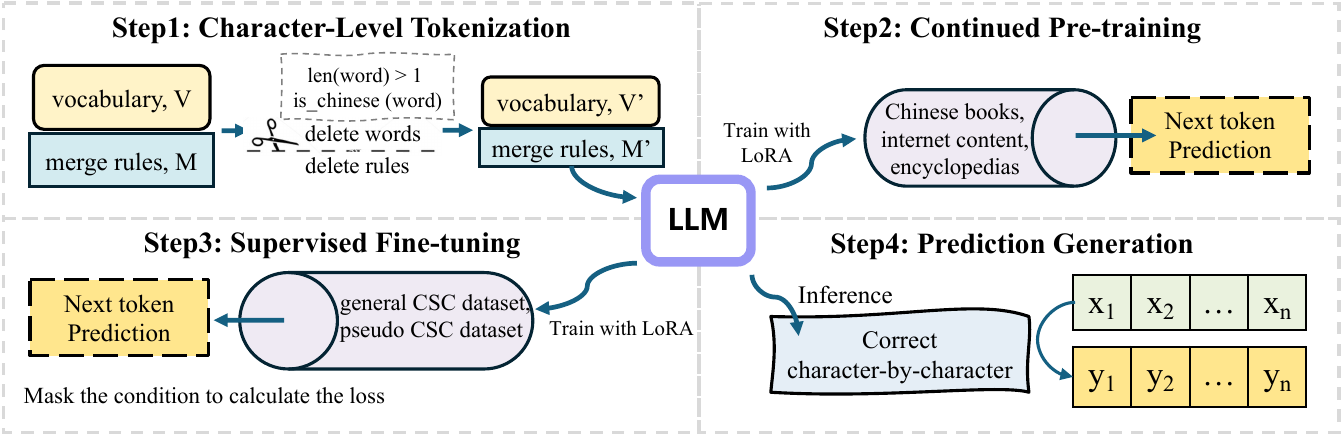}
    \caption{\label{fig:method}Overview of C-LLM. With an LLM (e.g., QWEN \cite{bai2023qwen}) as the core, the implementation process of C-LLM consists of multiple steps as illustrated in the figure.}
\end{figure*}

\subsection{Mixed Character-Word Tokenization}
By analyzing the tokenization used by the LLMs for CSC, we find that the current mixed character-word tokenization is the primary reason why LLMs struggle to meet the character-level length and phonetics constraints. Under this tokenization, sentences with spelling errors result in a character-to-word mapping that prevents LLM from establishing a clear character-level alignment. We analyze this issue through the following two scenarios  (see cases in Appendix~\ref{app:example}), where $x_{c_e}$ and $y_{c_e}$ denotes the erroneous character and the corresponding reference character, respectively, "$\Rightarrow$" denotes the correspondence between the tokens and characters:
\begin{align}
\label{eq:1} & x_{t_{i}} \Rightarrow \left\{x_{c_{e-1}} \right\}, x_{t_{i+1}} \Rightarrow \left\{x_{c_{e}}, x_{c_{e+1}} \right\} \\ 
\label{eq:2} & y_{t_i} \Rightarrow \left\{y_{c_{e-1}},y_{c_e},y_{c_{e+1}} \right\} 
\end{align}
(1) Comparing Equation~\ref{eq:1}\textasciitilde~\ref{eq:2}, the number of tokens in the source sentence does not match the reference sentence, resulting in multiple tokens corresponding to a single token. 
\begin{align}
\label{eq:3} & x_{t_{i}} \Rightarrow \left\{x_{c_{e-1}} \right\}, x_{t_{i+1}} \Rightarrow \left\{x_{c_e}, x_{c_{e+1}} \right\} \\
\label{eq:4} & y_{t_i} \Rightarrow \left\{y_{c_{e-1}},y_{c_e} \right\}, y_{t_{i+1}} \Rightarrow \left\{y_{c_{e+1}} \right\}
\end{align}
(2) In Equation~\ref{eq:3}\textasciitilde~\ref{eq:4}, even if the token counts are consistent, the characters may not align clearly due to erroneous characters and reference characters being placed in mismatched tokens. 
 
In both cases, LLM cannot directly map characters (e.g., $x_{c_e}$ -> $y_{c_e}$). This leads to three problems: (1) The inconsistency in the number of tokens between sentence pairs prevents LLM from learning the constraint of equal character length. (2) The unclear character correspondence hinders LLM from learning the constraint of similar character pronunciation. (3) The CSC task becomes more complex, involving numerous inference scenarios rather than character copying and replacement.

% In both cases, the mixed character-word tokenization complicates the direct alignment of $x_{c_e}$ and $y_{c_e}$, necessitating inference by the model to learn the correct character-level alignments. This transforms the CSC task into a semantic inference problem. Furthermore, inconsistencies in token counts and unclear character alignments hinder the model's ability to effectively learn character-level length and phonetic constraints.
However, in the CSC task, most correct characters in the source sentence can be directly copied during prediction, with only a small proportion of misspelled characters requiring replacement. Therefore, establishing a clear alignment between characters is crucial for this task. 

% In both cases, the mixed character-word tokenization complicates the direct alignment of $x_e$ and $y_e$, necessitating inference by the model to learn the correct mappings. 
% 然后说导致了3个问题：(1) 学不到长度相等 (2)学不到拼音相近 (3)任务复杂了，存在大量推理场景而不是复制和替换 ✅
% 整个这段的逻辑是：我们深度分析为什么当前mix-token不能建立字符级别的映射，然后说明 ✅

\section{Methodology}

% 围绕LLM（例如ChatGPT）为核心，HuggingGPT的实现流程由图中的四个阶段组成.

The CSC task requires a character-level mapping, necessitating character-by-character correction rather than token-by-token. Since current LLMs process sentences at the token level, mapping each character to a token can intuitively reduce the complexity of CSC for LLMs. Based on this concept, we propose C-LLM (as shown in Figure~\ref{fig:method}), a \textbf{L}arge \textbf{L}anguage \textbf{M}odel-based Chinese Spell \textbf{C}hecking method that learns to check errors \textbf{C}haracter by character. This approach consists of three main steps, as detailed below.

% 这个逻辑不通： 存在大量的字符复制和小部分的字符替换，需要一个字一个字的检查，而不是一个token一个token的检查
% 应该是csc任务的要求是字符级别的一一映射，所以需要一个字一个字的检查，而不是一个token一个token的检查 ✅
% This approach learns to detect errors on a character-by-character basis, as detailed below.  这里说一下分成三个步骤，具体如下。 ✅

% 根据上述分析，CSC任务本身存在大量的字符复制和小部分的字符替换，需要一个字一个字的检查，而不是一个token一个token的检查。鉴于目前LLM都是以分词后的token为基本单位来理解句子，因此直觉上，把每个字符映射到一个token上可以大大降低llm理解CSC任务的难度.基于这个思想，我们提出了c-llm（如图a所示），一种基于大语言模型的汉语拼写检查方法，可以逐字逐字地学习检查错误，具体介绍如下。

\subsection{Character-Level Tokenization}
\begin{table*}[!ht]
    \tiny
    \centering
    \resizebox{0.95\linewidth}{!}{
    \begin{tabular}{l|cccccc|c}
    \hline
        Models  & Government &  Movie  & General  & Game  & Tech  & Finance &  Avg \\ \hline
        Original-7B & 8.84  &  50.27 & 12.57 & 37.19 & 28.16 & 10.18 & 24.53   \\
        Char-7B  & 164.12  &  931.99 & 170.02 & 641.76 & 560.99 & 120.99 & 431.65   \\ 
        Char-PT-7B & 11.80  &  64.48 & 14.92 & 48.90 & 34.99 & 11.89 & 31.16  \\ \hline
        Original-14B & 8.25  & 46.67  & 11.75  & 34.60  & 25.57  & 9.49  & 22.72   \\
        Char-14B  & 131.31  & 758.01  & 130.71  & 506.21  & 410.33  & 95.40  & 338.66   \\ 
        Char-PT-14B & 10.51  & 58.76  & 14.13  & 44.04  & 32.20  & 11.63  & 28.55  \\ \hline
    \end{tabular}
    }
    \caption{\label{tab:qwen_ppl}The perplexity of LLMs (e.g., QWEN1.5-14B and QWEN1.5-7B) were evaluated using the Chinese domain modeling dataset (from Skywork \cite{wei2023skywork}). "Original" refers to the original LLMs, "Char" denotes LLMs with character-level tokenization, and "Char-PT" indicates the model that was further pre-trained.}
\end{table*}
% The table presents results specifically for QWEN14b, while perplexity for other model sizes are available in the Appendix.

% 在Skywork的中文领域建模能力评测数据集上测试各个LLM困惑度的测试结果。表中只展示了QWEN14b下的结果，其他模型规模的困惑度见附录。“Ori”表示原始LLM，“Char”表示约束词表后的模型，“char-PT”表示在其基础上继续预训练后的模型。

% 1. 裁剪词表
The vocabulary of LLMs is typically multilingual. However, since CSC primarily addresses errors in Chinese, we only focus on the Chinese portion of the vocabulary. As shown in Equations~\ref{eq:1}$\sim$\ref{eq:4}, LLMs often map multiple characters to a single token during tokenization, complicating the CSC task by preventing a direct alignment between characters. To mitigate this issue, we construct character-level tokenization to ensure that each Chinese character is mapped to a single token. This approach facilitates a clear alignment between characters in the tokenized sentences, as represented by the following equation:
\begin{align}
\label{eq:5} & x_{t_i}\!\Rightarrow\!\left\{x_{c_{e-1}} \right\}, x_{t_{i+1}}\!\Rightarrow\!\left\{x_{c_e}\right\}, x_{t_{i+2}}\!\Rightarrow\!\left\{x_{c_{e+1}} \right\} \\ 
\label{eq:6} & y_{t_i}\!\Rightarrow\!\left\{y_{c_{e-1}} \right\}, y_{t_{i+1}}\!\Rightarrow\!\left\{y_{c_e}\right\} , y_{t_{i+2}}\!\Rightarrow\!\left\{y_{c_{e+1}} \right\} 
\end{align}

Specifically, the approach for constructing the
character-level tokenization of LLM (e.g., QWEN \cite{bai2023qwen}), is detailed in Algorithm~\ref{alg:vocab_filter}. For the BPE \cite{gage1994new} tokenization, we refine the vocabulary and the merging rules. With the new vocabulary, the model is unable to recognize words composed of multiple Chinese characters, resulting in each Chinese character being mapped to a separate token according to the revised merging rules. Experimental results indicate that the new vocabulary size is reduced to 89.2\% of the original.

\renewcommand{\algorithmicrequire}{ \textbf{Input:}} 
\renewcommand{\algorithmicensure}{ \textbf{Output:}} 
\begin{algorithm}[!ht]
\caption{Methods for Constructing Our Character-Level Tokenization.}
\label{alg:vocab_filter}
\begin{algorithmic}[1] %这个1 表示每一行都显示数字
\REQUIRE ~~\\ %算法的输入参数：Input
    The vocabulary of LLMs, $V$; The merge rules applied during tokenization, $M$.
\ENSURE ~~\\ %算法的输出：Output
    The updated vocabulary $V^{'}$ and merge rules $M^{'}$ for the LLMs;
\STATE Initialization: The list of word $D_w$ and the list of merging rules $D_m$ to be filtered.
\FOR{$word$ in $V$}
\IF{len($word$) > 1 and $word$ is chinese string}
    \STATE add $word$ in $D_w$; update $D_w$;
    \ENDIF
    \ENDFOR
\FOR{$merge\_rule$ in $M$}
\STATE $a, b = merge\_rule$[0], $merge\_rule$[1]
\IF{decode($a + b$) in $D_w$ or decode($a$) in $D_w$ or decode($v$) in $D_w$}
    \STATE add $merge\_rule$ in $D_m$; update $D_m$;
    \ENDIF
    \ENDFOR
\STATE Update $V$ and $M$ by removing the words and merge rules recorded in $D_w$ and $D_m$, resulting in $V^{'}$ and $M^{'}$.
\RETURN $V^{'}$ and $M^{'}$.
\STATE Update the model's input and output embedding according to the new vocabulary $V^{'}$.
\end{algorithmic}
\end{algorithm}

\subsection{Continued Pre-training}
\label{sec:ppl}
% 2. 继续预训练
To mitigate the potential impact on the LLM's language modeling ability due to vocabulary constraints, we continued pre-training LLM (based on QWEN \cite{bai2023qwen}) to adapt it to the new vocabulary. Specifically, we performed continued pre-training with LoRA \cite{hu2021lora} on the Chinese open-source pre-training dataset provided by Tigerbot \cite{chen2023tigerbot}, which includes Chinese books, internet content, and encyclopedias. The training data comprised approximately 19B tokens, but we trained for 30,000 steps, covering about 2B tokens. More implementation details are provided in the Appendix~\ref{sec:appendix param}.

% 这个loss公式可以不写，比较简单，而且这里引入了新的记号 t_{i}，跟前面一起看会比较乱。 ✅

To evaluate the impact of the character-level tokenization and continued pre-training on the LLM's language modeling ability, we measure the perplexity of LLMs using the Chinese domain modeling competency assessment dataset from Skywork \cite{wei2023skywork}. As shown in Table~\ref{tab:qwen_ppl}, the perplexity increased significantly after applying character-level tokenization, indicating a substantial impact on language modeling ability. However, this effect was mitigated after continued pre-training, bringing the language modeling ability close to that of the original LLM. This demonstrates that the model effectively adapted to the new vocabulary. 

\subsection{Supervised Fine-tuning}
% 3. 纠错数据SFT
% 继续预训练后模型只学习了语言的通用特征，并不能理解CSC任务要做什么，因此还需要进行有监督的微调，让LLM学习到与CSC相关的特征。微调的细节见附录。
After continue pre-training, LLM only learns general language features and does not understand the specific requirements of the CSC. Therefore, supervised fine-tuning is necessary for the LLM to learn the CSC task. We utilize LoRA \cite{hu2021lora} for the fine-tuning. The training loss is defined as follows and the implementation details are provided in Appendix~\ref{sec:appendix param} and Section~\ref{sec:experiment}.
\begin{align}
\mathcal{L}(\mathcal{T})\!=\!\sum_{i=1}^{N}\!\log (\mathbb{P}\left(Y^{'}_c \!\mid I, X_c\right.)) \label{eq:8} 
\end{align}
where loss is calculated as the conditional probability of the predicted sentence $Y^{'}_c$ given the task description of the CSC $I$ and source sentence $X_c$.

% yt_{i} => y_{t_{i}}, log(P(Y|I,X)), I表示CSC的任务描述，X表示输入原文 这样就行 ✅

\section{Experiments}
\label{sec:experiment}
In this section, we present the details of fine-tuning and the evaluation results of models on the two CSC benchmarks: the general dataset CSCD-NS and the multi-domain dataset LEMON.

\subsection{Fine-tuning Datasets and Metrics}
\textbf{Datasets} Previous studies \cite{liu-etal-2021-plome,xu-etal-2021-read} chose SIGHAN \cite{wu2013chinese,yu2014overview,tseng2015introduction} as the benchmark. However, an increasing number of studies \cite{hu2022CSCD,yin2023comprehensive,li2022improving} have identified numerous issues with this dataset, such as semantically incoherent and annotation errors. Consequently, in our study, we chose two new CSC benchmarks, namely CSCD-NS and LEMON: (1) CSCD-NS \cite{hu2022CSCD}: CSCD-NS superior in quality to SIGHAN, is the first CSC dataset where the primary source of character errors stems from pinyin input methods, containing a significant amount of homophonic and word-level errors. (2) LEMON \cite{wu2023rethinking}: LEMON is a novel, large-scale, multi-domain CSC dataset featuring various real-world spelling errors. It spans seven different sub-domains, including game (GAM), encyclopedia (ENC), contract (COT), medical care (MEC), car (CAR), novel (NOV), and news (NEW), typically testing the model's domain correction capabilities in a zero-shot setting. Appendix~\ref{app:data_stat} shows the data statistics.

Following the fine-tuning approach of previous work \cite{li2022improving,liang2023disentangled}, we combined the training data from CSCD-NS and 271K pseudo-data generated by ASR or OCR (denoted as Wang271K) \cite{wang-etal-2018-hybrid} as our training set. The validation data from CSCD-NS was used as our validation set, and we test the models on the CSCD-NS test data and LEMON, respectively.

\textbf{Evaluation Metrics} 
We report sentence-level and character-level precision, recall, and F1 scores to evaluate different models. These metrics are reported separately for detection and correction tasks. We calculate metrics using the script from CSCD-NS \cite{hu2022CSCD}. For predictions from LLMs that do not match the source sentence length, we first employ ChERRANT \cite{zhang-etal-2022-mucgec} to extract non-equal length operations, then replace these with the source before calculating the metrics.

\subsection{Baselines}

% baseline
We use the following CSC models for comparison.
\textbf{BERT-style models}. (1) BERT \cite{devlin-etal-2019-bert}: BERT approaches CSC as a sequence labeling task, encoding the input sentence and employing a classifier to select the appropriate characters from the vocabulary. (2) Soft-Masked BERT (SMBERT) \cite{zhang-etal-2020-spelling}: SMBERT composed of a detection and correction network, enhances BERT's error detection capabilities. (3) SCOPE \cite{li2022improving}: SCOPE incorporates an auxiliary pronunciation prediction task with an adaptive task weighting scheme to improve CSC performance.

For the selection of LLMs, we carry out a series of experiments using QWEN1.5 \cite{bai2023qwen}. As one of the most potent open-source LLMs in China, QWEN exhibits robust Chinese processing capabilities and has released model parameters of multiple scales. We evaluate the performance of LLMs under the following two settings, and the prompts for LLMs are detailed in the Appendix~\ref{app:prompt_cscd}.

\textbf{Fine-tuned LLM} (LLM-SFT): The original LLMs (Original), the LLMs with character-level tokenization (Char), and the further pre-trained character-level LLMs (Char-PT) are each fine-tuned on the aforementioned dataset.

\textbf{LLM with In-Context Learning} (LLM-ICL): The original LLMs (Original), ChatGPT and GPT-4 are adapted to perform the CSC task using prompts.

\begin{table*}[htbp]
    \centering
    \resizebox{\linewidth}{!}{
    \begin{tabular}{l|cccccccc|c}
    \hline
        Models & CAR & COT & ENC & GAM & MEC & NEW & NOV & CSCD-NS & Avg \\ \hline
        BERT \cite{devlin-etal-2019-bert} & 46.87  & 52.61  & 45.74  & 23.41  & 42.73  & 46.63  & 32.35  & 65.49  & 44.48   \\ 
        SMBERT \cite{zhang-etal-2020-spelling} & 49.91  & 54.85  & 49.33  & 26.18  & 46.91  & 49.16  & 34.56  & 67.22  & 47.26   \\ 
        SCOPE \cite{li2022improving} & 50.71  & 54.89  & 45.23  & 24.74  & 44.44  & 48.72  & 33.17  & 71.70  & 46.70  \\ \hline
        ChatGPT & 44.88  & 57.11  & 51.46  & 28.78  & 49.85  & 44.40  & 31.77  & 52.50  & 45.09   \\ 
        GPT-4 \cite{achiam2023gpt} & 54.44  & 62.82  & 55.12  & 36.27  & 56.36  & 56.09  & 45.64  & 54.41  & 52.64   \\ \hline
        Original-ICL-7B & 21.48  & 37.33  & 33.38  & 22.12  & 27.82  & 23.95  & 19.22  & 19.10  & 25.55  \\
        Original-SFT-7B & 53.38  & 56.55  & 54.44  & 37.33  & 59.21  & 58.96  & 39.12  & 68.66  & 53.46   \\
        Char-SFT-7B & 52.10  & 57.02  & 52.55  & \textbf{39.00}  & 59.85  & 59.01  & 40.34  & 70.41  & 53.78   \\ 
        Char-PT-SFT-7B (C-LLM) & 53.87  & 58.04  & 54.57  & 37.43  & 61.16  & 60.07  & 41.42  & 71.64  & 54.77   \\ 
        \hline
        Original-ICL-14B & 36.75  & 46.72  & 46.92  & 25.15  & 42.48  & 40.40  & 31.27  & 41.09  & 38.85  \\
        Original-SFT-14B & 54.56  & 56.82  & 53.44  & 32.59  & 58.89  & 63.32  & 40.58  & 72.63  & 54.10 \\
        Char-SFT-14B  & 55.36  & 59.11  & 54.30  & 37.21  & 60.43  & \textbf{65.28}  & 42.33  & 72.78  & 55.85   \\ 
        Char-PT-SFT-14B (C-LLM) & \textbf{57.54}  & \textbf{60.40}  & \textbf{56.48}  & 38.02  & \textbf{65.31}  & 64.49  & \textbf{43.92}  & \textbf{73.80}  & \textbf{57.49}   \\ 
        \hline
    \end{tabular}
    }
    \caption{\label{tab:main_result}Overall performance (\%) of C-LLM and baseline models, are presented as character-level correction F1 scores. The best results are highlighted in bold. All the results of the BERT-style models are reproduced by us.}
\end{table*}
% C-LLM和基线微调后在通用数据CSCD和领域数据LEMON上的字符级纠错F1分数。最好的结果以粗体显示。其中基线结果均由我们自己复现。

\subsection{Main Results}
The main results on the CSCD-NS and LEMON test sets are presented in Table~\ref{tab:main_result}, revealing several observations: (1) The model's error correction performance with prompts is suboptimal. Even with GPT-4, achieving satisfactory results is challenging. However, supervised fine-tuning significantly improves performance, emphasizing its importance. (2) Compared to C-LLM, LLMs without continued pre-training (Char-SFT) show a decline in average performance, highlighting the necessity of continued pre-training for better adaptation to new vocabulary and improved performance. This is also evident in the perplexity comparison in Section~\ref{sec:ppl}. (3) In domain-specific data, the concise nature of news language in the NEW dataset and the idiomatic expressions in the GAM dataset make models with continued pre-training more prone to incorrect corrections. (4) The original LLM outperforms BERT-style models in error correction, indicating that LLMs have an advantage over BERT-style models in CSC tasks, especially in vertical domains, consistent with the insights in Section~\ref{sec:related work}. (5) C-LLM demonstrates superior error correction performance in both general and vertical domains compared to BERT-style models and the original LLM, achieving state-of-the-art performance. This confirms the effectiveness of character-level error correction.

% 在领域数据中，由于NEW中新闻语言的简洁性以及 GAM中游戏语言的非正式性，经过持续预训练的大模型更容易错误地纠正。
% Origin-SFT-14B已经明显比BERT好了，这要单独说一下，LLM在CSC任务上还是有优势的，尤其垂直领域，呼应related work中bert参数不足的表述。 ✅
% 其实就几个观点: (1) SFT很重要，ICL很难work，及时GPT4也不行，SFT之后大模型效果提升明显，也比BERT强了 (2) 裁剪词表后pretrain很重要 (3) C-LLM （字符级别+PT+SFT）获得了sota ✅

% scaling趋势
\begin{figure*}[!ht]
    \centering
    \includegraphics[width=0.95\linewidth]{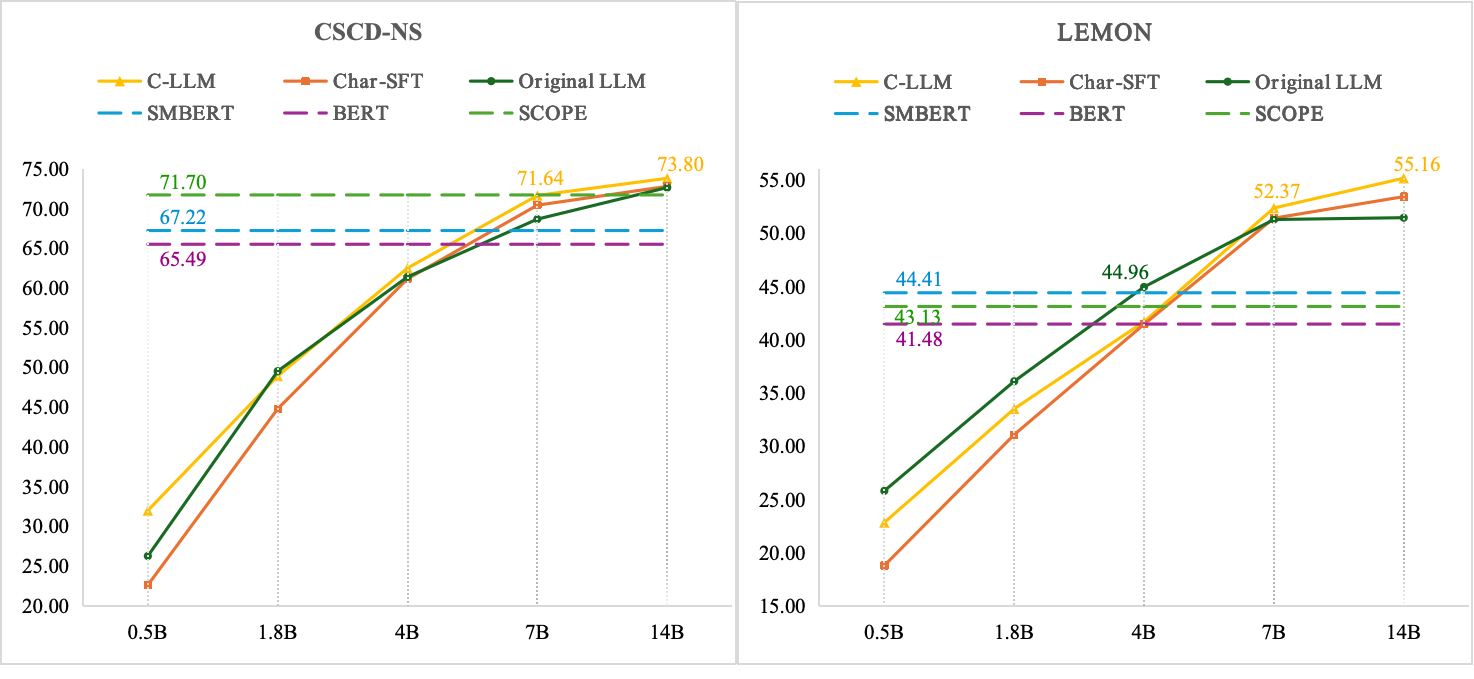}
    \caption{\label{fig:scaling}The trend of character-level correction F1 scores for C-LLM (based on QWEN) across various parameter. Results are presented for both CSCD-NS and LEMON datasets.}
\end{figure*}
% C-LLM（基于QWEN）在各个参数规模下的字符级纠错F1分数趋势。图中分别展示了在CSCD-NS和LEMON上的结果。

\section{Analysis and Discussion}
In this section, we further analyze and discuss our model from both quantitative and qualitative perspectives.
% 在本节中，我们进一步从各个角度定量和定性地分析和讨论我们的模型。
% Specifically, we focus on three main aspects: parameter size, character-level constraints, and inference speed.

\subsection{Scaling Trends}

To further investigate the impact of model size on correction performance for LLMs, we also conduct experiments under 4B, 1.8B, and 0.5B parameters, while keeping the fine-tuning dataset and training hyperparameters consistent. As shown in Figure~\ref{fig:scaling}, the correction performance of the LLMs decreases on both test sets as the parameter size reduces.

Comparing C-LLM with BERT-style models, C-LLM outperforms BERT-style models at both 14B and 7B parameter sizes on the CSCD-NS and LEMON, particularly excelling in vertical domain tasks. However, smaller models exhibit weaker performance. We speculate that despite the simplification of the CSC through character-level tokenization, smaller models still struggle to understand the task adequately, resulting in poor performance.

% We believe that  尽量不要用这个表达，这个要给出实验证据的。就我们猜测。。。 ✅

Comparing C-LLM with the original LLM, C-LLM consistently outperforms the original LLM across various parameter sizes on the CSCD-NS dataset, although the performance gap narrows at 1.8B. This indicates that C-LLM has superior error correction capabilities compared to the original LLM. However, on the LEMON dataset, C-LLM underperforms the original LLM at sizes of 4B and smaller. We attribute this to the substantial amount of domain-specific data included in the pre-training of original LLM \cite{bai2023qwen}, whereas our continued pre-training for C-LLM only includes general Chinese data. This may lead to the forgetting of some domain knowledge in LLM. Larger C-LLM models (14B and 7B) suffer less from this forgetting due to their larger parameter sizes. Despite some domain knowledge being forgotten, the character-level correction approach allows larger C-LLM models to achieve better performance, while smaller models are more affected by knowledge forgetting, resulting in poorer performance.

Comparing C-LLM with Char-SFT, Char-SFT consistently underperforms C-LLM across both datasets and all model sizes. This underscores the importance of continued pre-training, which enables the model to better adapt to new vocabulary and achieve improved performance.

\subsection{Analysis of Length and Phonetic}

% 长度和拼音角度分析
\begin{table}[htbp]
    \centering
    \scalebox{0.75}{
    \begin{tabular}{l|c|cc}
    \hline
        Models & Equal-length & Non-homophon & Ratio \\ \hline
        Original-ICL-14B &  22.86\% & 53.70\% & 84.42\% \\
        Original-SFT-14B & 96.92\% & 8.63\% & 38.52\% \\
        Char-PT-SFT-14B &  99.78\% & 3.83\% & 18.43\% \\ \hline
        Target & 100\% & 1.74\% & / \\ \hline
    \end{tabular}
    }
    \caption{\label{tab:len_stat}Statistical results from the length and phonetic perspective, using the 14B models as an example. "Target" refers to the reference sentences in the test set. "Ratio" indicates the ratio of non-homophone characters in incorrect predictions.}
\end{table}

% Perspective & C-LLM & Original-SFT & Original-ICL  \\ \hline
% Char-to-token & 98.19\% & 56.48\% & / \\
% Token-level & 98.84\% & 80.54\% & / \\
% Character-level & 99.78\% & 96.92\% & 22.86\% \\ 

% Models & Predicted Char & Wrong Char & Ratio of Errors  \\ \hline
% Original-ICL & 53.70\% & 99\% & 84.42\% \\
% Original-SFT & 8.63\% & 88.95\% & 38.52\%  \\ 
% C-LLM & 3.83\% & 86.07\% & 18.43\% \\ \hline

% \textbf{C-LLM alleviates issues related to character-level length constraints.} To evaluate whether C-LLM effectively addresses the issue of LLMs failing to meet character-level length constraints, we analyzed from following perspectives, with results presented in Table~\ref{tab:len_stat}.

\textbf{C-LLM alleviates issues related to character-level length constraints.} To evaluate the effectiveness of Char-PT-SFT (C-LLM) in addressing character-level length constraints, we select sentence pairs from the CSCD-NS test set. These pairs exhibit tokenization discrepancies between the source and reference sentences, highlighting character-to-word mapping issues. By comparing the model's predictions on these sentence pairs to see if it maintains the same number of characters as the source sentence, we can better assess its understanding of character-level length. As shown in Table~\ref{tab:len_stat}, Original-SFT increases the proportion of predictions maintaining the character-level length to 96.92\% compared to Original-ICL, indicating that fine-tuning helps LLMs adhere to character-level length constraints.

% C-LLM ensures character-level length consistency. 没有100%，不能说ensures，只能说 克服/极大缓解 ✅

% (1) Token-Level: Our analysis shows that 98.19\% of the tokens generated by C-LLM correspond one-to-one with Chinese characters. This results in approximately 18\% more sentences where the token count of the source sentence matches that of the reference, compared to the original LLM.

Under C-LLM, the consistency in character-level length further improves to 99.78\%. This finding demonstrates that the one-to-one correspondence between tokens and Chinese characters enables LLMs to more easily generate sentences that meet character-level length constraints, resulting in superior performance.
% For the original LLM, correctly predicting the character-to-word relationship is challenging and often leads to predictions that do not meet character length constraints.

% 这里叙述比较复杂了，直接给出ICL，origin-SFT，C-LLM，标准target（100%）， 字符级长度一致的占比就好（一个表），有直观的对比。  ✅
% 说明sft很重要，拆分char能进一步继续提升 ✅

\textbf{C-LLM can reduce phonologically dissimilar predictions.} We calculate the proportion of non-homophonic characters among all predicted characters and the proportion of non-homophonic errors among all incorrect predicted characters in the CSCD-NS test set. As shown in Table~\ref{tab:len_stat}, Original-ICL produces more than half of the non-homophonic errors, with the majority of its incorrect predictions being non-homophonic errors. In contrast, Original-SFT significantly reduced both proportions, indicating that supervised fine-tuning helps the LLMs maintain phonetic constraints. 

C-LLM generates fewer non-homophonic prediction errors, reducing the proportion of non-homophonic errors among total prediction errors by approximately 20\% compared to Original-SFT. This suggests that although C-LLM still produces some non-homophonic predictions, the impact of these errors on LLMs' correction performance has been greatly diminished.

% 这里叙述比较复杂了，直接给出ICL，origin-SFT，C-LLM，标准target， 拼音不一致的占比，有直观的对比。 ✅

% \begin{table}[htbp]
%     \centering
%     \resizebox{0.48\textwidth}{!}{
%     \begin{tabular}{l|ccc}
%     \hline
%         ~ & Original-SFT & C-LLM  \\ \hline
%         Non-homophon Predict &  8.63\% & 3.83\% \\
%         Ratio of Wrong Predict &  38.52\% & 18.43\% \\ \hline
%     \end{tabular}
%     }
%     \caption{\label{tab:pinyin}Statistical Results for non-homophone predicted characters (under 14B model parameters).}
% \end{table}
% Models & Predicted Char & Wrong Char & Ratio of Errors  \\ \hline
% Original-ICL & 53.70\% & 99\% & 84.42\% \\
% Original-SFT & 8.63\% & 88.95\% & 38.52\%  \\ 
% C-LLM & 3.83\% & 86.07\% & 18.43\% \\ \hline

\subsection{Inference Speed Analysis}

\begin{table}[tbp]
    \centering
    \resizebox{0.49\textwidth}{!}{
    \begin{tabular}{l|cccc}
    \hline
        Models & \#Tokens & \#Characters & AR & Time (s) \\ \hline 
        Original-SFT-7B & 83530 & 128676  & 86.50\% & 2028.77  \\
        Char-PT-SFT-7B & 127057  & 128801  & 93.88\% & 2481.97  \\
        \hline
    \end{tabular}
    }
    \caption{\label{tab:sample} Analysis of Inference Speed. "AR" indicates the acceptance rate generated by draft model.}
\end{table}

Using a character-level tokenizer can decrease the model's inference speed. In this study, we perform a quantitative analysis of this impact by employing speculative decoding \cite{chen2023accelerating}. Our evaluation uses samples containing spelling errors from the CSCD-NS test set. The target model has 7B parameters, while the draft model has 1.8B parameters, with draft tokens set to 4. Specifically, to test the speculative decoding capability of Original-SFT-7B, we use Original-SFT-1.8B as the draft model. For Char-PT-SFT-7B, we use Char-PT-SFT-1.8B as the draft model.

As shown in Table~\ref{tab:sample}, under Char-PT-SFT-7B, the number of decoded tokens increased by 52\% compared to Original-SFT-7B, but the overall time consumption only increased by 22.33\%. This is because the task complexity was reduced by Char-PT-SFT-7B, leading to a higher acceptance rate for speculative decoding compared to original LLM.

% 最前面加一句，tokenizer变成char级别会降低模型推理的速度，我们在这里进行量化的分析。我们采用投机解码的方式来加速推理。测试数据为CSCD的测试集，含有错别字的样本。
% 这个表不用这么复杂，就对比 总token数量，总char数量，接受率，总耗时 ✅
% 方法应是CLLM，Original-SFT ✅
% "Ratio" indicates the acceptance rate of tokens generated by draft model. => "AR" indicates the acceptance rate generated by draft model. ✅

\section{Conclusion}

This paper indicates that LLMs fail to meet the Chinese character-level constraints of the CSC task, namely equal length and phonetic similarity, which hinders their correction performance. We find that the root cause lies in the granularity of tokenization, which mixes characters and words, making it difficult to satisfy these character-level constraints. To address this issue, we propose C-LLM, which establishes mappings between Chinese characters, enabling the model to learn correction relationships and phonetic similarities. This approach simplifies the CSC task to character replication and substitution. Experimental results demonstrate that C-LLM outperforms previous methods on both general and multi-domain benchmarks, achieving state-of-the-art performance.

% notes 这个词不好 ✅
% find that the root cause 不太地道，我们揭示了 

\section{Limitations}
Our work has three main limitations. First, our method is specifically designed for Chinese spelling checking and may not effectively address sentences with English errors, as we did not process English words in the vocabulary. Second, our model has room for improvement, especially in handling new and trending words, which may require integrating methods such as RAG. Finally, our model's inference time is longer compared to the original model, indicating a need for further optimization for practical applications.
% 我们的局限性主要有三点。首先因为主要针对的是中文拼写纠错这个任务，我们的工作重点是中文，对于存在英文错误的句子，由于没有对词表中的英文进行处理，我们的方法可能不会带来更多性能的提升。其次，我们的模型仍然有改进的空间，对于一些新热词可能需要结合RAG的方法来做纠错。最后，我们的模型推理时间相比原来的模型更长，在应用时可能需要进一步优化。

\section*{Acknowledgments}
We would like to thank the National Science and Technology Major Project (2022ZD0115801) for the generous grant.

\bibliography{anthology,custom}

\appendix
\clearpage
\section{Appendix}
\label{sec:appendix}

\subsection{Implementation Details}
\label{sec:appendix param}

\textbf{Hyparameters of Continued Pre-training} Our experiments are conducted on eight NVIDIA A100-SXM4-40GB GPUs. We provide a overview of the hyperparameter settings used in continued pre-training with LoRA \cite{hu2021lora}, as illustrated in Table~\ref{tab:hyparam_pt}. Our implementation is based on Huggingface’s Transformers \cite{wolf-etal-2020-transformers} in PyTorch.

% 我们

\begin{table}[htbp]
    \centering
    \begin{tabular}{l|c}
    \hline
        Configurations & Values \\ \hline
        learning\_rate  & 1e-5 \\
        batch\_size & 128 \\
        adam\_beta1 & 0.9 \\
        adam\_beta2 & 0.999 \\
        adam\_epsilon & 1e-8 \\
        tokens/batch  & $2^{16}$ \\  
        steps & 30000 \\
        lora\_r & 16 \\
        lora\_alpha & 32 \\
        lora\_dropout & 0.1 \\ \hline
    \end{tabular}
    \caption{Hyparameters used in continued pre-training.}
    \label{tab:hyparam_pt}
\end{table}

% Batch\_size for pre-training & 1 \\ 改成 tokens/batch => 512 * 128 = 2^16,  gradient\_accumulation\_steps 删除这一项，max\_steps => steps, 30000, 最好在正文中也说一下，我们训练了3w步，约2B tokens。前文中说tiger19B，但是我们只训练了2B。  ✅

\textbf{Hyparameters of Supervised Fine-tuning} Our experiments are conducted on eight NVIDIA A100-SXM4-40GB GPUs. We also provide the overview of the hyperparameter settings used in fine-tuning with LoRA \cite{hu2021lora}, as illustrated in Table~\ref{tab:hyparam_ft}.

\begin{table}[htbp]
    \centering
    \begin{tabular}{l|c}
    \hline
        Configurations & Values \\ \hline
        learning\_rate  & 1e-4 \\
        batch\_size & 32 \\
        adam\_beta1 & 0.9 \\
        adam\_beta2 & 0.999 \\
        adam\_epsilon & 1e-8 \\
        num\_train\_epochs & 10 \\
        lora\_r & 16 \\
        lora\_alpha & 32 \\
        lora\_dropout & 0.1 \\ \hline
    \end{tabular}
    \caption{Hyparameters used in fine-tuning.}
    \label{tab:hyparam_ft}
\end{table}

\begin{CJK*}{UTF8}{gbsn} 
\begin{table*}[!ht]
    \centering
    % \resizebox{0.5\textwidth}{!}{
    \renewcommand{\arraystretch}{1.1}
    \begin{tabular}{l|p{13cm}}
    \hline
        \textit{\textbf{Case1}} & Sentence pair with \textbf{inconsistent} token counts \\ \hline
        Source 1 & \textcolor{red}{胜}/ 名/ 的/ 酒店 \quad \textcolor{red}{Winning} Hotels \\
        Reference 1 & \textcolor{blue}{著}名的/ 酒店 \quad\textcolor{blue}{Famous} Hotels\\ \hline
        
        \textit{\textbf{Case2}} & Sentence pairs with \textbf{consistent} token counts \\ \hline
        Source 1 & 高\textcolor{red}{达}/ 的/ 公众/ 形象 \quad \textcolor{red}{Gundam's} public image \\
        Reference 1 & 高/ \textcolor{blue}{大}的/ 公众/ 形象 \quad \textcolor{blue}{Tall} public image  \\
        Source 2 & 游戏/ 展开/ \textcolor{red}{目}前/ 一天 \quad  The game unfolds for the \textcolor{red}{current} day \\
        Reference 2 & 游戏/ 展/ 开\textcolor{blue}{幕}/ 前一天  \quad The day before the game \textcolor{blue}{begins} \\ \hline
    \end{tabular}
    % }
    \caption{Examples illustrating the tokenization mismatches in two scenarios. '/' indicates participle position.}
    \label{tab:example}
\end{table*}
\end{CJK*}

\subsection{Examples for illustration}
\label{app:example}
During model training, the mapping between source tokens and reference tokens is learned. The Table~\ref{tab:example} presents examples illustrating the mismatch between source and reference tokenization in two scenarios, using sentences containing a single error as examples:

\begin{CJK*}{UTF8}{gbsn}
(1) Case1 corresponds to Equations~\ref{eq:1}\textasciitilde~\ref{eq:2} in the paper, where the number of tokens in the source sentence does not match the reference sentence, resulting in multiple tokens mapping to a single token (e.g., "胜(win)", "名(name)", "的(of)"->"著名的(famous)").

(2) Case2 corresponds to Equations~\ref{eq:3}\textasciitilde~\ref{eq:4}, where the token counts are consistent, but the characters may not align clearly due to erroneous and reference characters being placed in mismatched tokens (e.g., The tokens where "达(reach)" and "大(big)" are located are not aligned). However, even if the characters can be placed in matched tokens (e.g. "目前(present)"->"开幕(open)"), the semantic correspondence between tokens may be disrupted due to improper tokenization.

The examples above fail to establish a clear character-level mapping, requiring the model to deduce implicit character alignment (e.g., "胜(win)"->"著(write)", "达(reach)"->"大(big)", "目(eye)"->"幕(screen)"). This complicates the CSC by turning it into a semantic inference problem, thereby hindering the model’s ability to effectively learn character-level length and phonetic constraints.

\end{CJK*}

% Batch\_size for pre-training & 1 \\ gradient\_accumulation\_steps & 32 \\ ， 整体写batch size：32就行 ✅
\subsection{Prompts Setting}
\label{app:prompt_cscd}
Table~\ref{tab:prompt_cscd} presents the prompts used to evaluate the error correction performance of the fine-tuned LLM, along with the few-shot prompts for ChatGPT, GPT-4 and Original-ICL. The few-shot prompt consists of 10 examples: 5 sentence pairs without typos and 5 with typos. These positive and negative examples are randomly selected from CSCD-NS, and their positions within the prompt are also randomized.
% 表1中展示了微调后的LLM在测试纠错性能时使用的prompt，以及用于ChatGPT和GPT-4的few-shot prompt。其中few-shot prompt中设置了10个样例，包含5个没错别字的句子对和5个含错别字的句子对，正负样例是随机从CSCD-NS中挑选， 正负样本所在的位置也是随机的。

\begin{CJK*}{UTF8}{gbsn} 
\begin{table*}[htbp]
    \centering
    % \resizebox{0.5\textwidth}{!}{
    \renewcommand{\arraystretch}{1.1}
    \begin{tabular}{l|p{10cm}}
    \hline
        Models & Prompts \\ \hline
        \multirow{2}{*}{Fine-tuned LLM} & 任务: 纠错文本, 输入: "原句", 输出: \\ 
        &  (Task: Correct the text, Input: $\left\{source\_sentence\right\}$, Output:)  \\ \hline
        \multirow{2}{*}{ChatGPT, GPT-4, Original-ICL} & 纠正句子中的错别字，并返回纠正后的句子。 (Identify and correct the spelling errors in the sentence, then provide the corrected version.) \\
        & $ \left\{sentence1\right\}\!=>\!\left\{reference\_sentence1\right\} ... \left\{sentence10\right\}=>\left\{reference\_sentence10\right\} => \left\{source\_sentence\right\} => $ \\ \hline
    \end{tabular}
    % }
    \caption{Prompts used for testing.}
    \label{tab:prompt_cscd}
\end{table*}
\end{CJK*}

% ICL的例子应该有换行吧？ Original-ICL是啥，表3中就是ChatGPT和GPT4

\subsection{Data Statistics}
\label{app:data_stat}
The statistical results for the Wang271K, CSCD-NS and LEMON datasets are presented in Table~\ref{tab:data_stat}. The LEMON spans seven different sub-domains, including game (GAM), encyclopedia (ENC), contract (COT), medical care (MEC), car (CAR), novel (NOV), and news (NEW). To better evaluate model performance, we filtered out sentences from the LEMON dataset where the source and reference sentences had unequal character-level lengths or where the source sentence exceeded 1000 characters.
% 关于CSCD-NS和LEMON数据的统计结果见表a。其中为更好地测试模型性能，我们筛选过滤掉LEMON数据集中原句和参考句长度不等，或原句长度大于1000的句子。

\begin{table*}[!ht]
    \centering
    \begin{tabular}{l|cccc}
    \hline
        \textbf{\textit{Train}} & \#Sent & \#Errors & \#Phonetically Similar Errors & Avg.Length \\ \hline
        CSCD-NS & 29,999 & 15,142 & 14,804 & 57.39 \\
        Wang271K & 271,329 & 381,962 & 157,907 & 44.4 \\ \hline
        \textbf{\textit{Dev}} & \#Sent & \#Errors & \#Phonetically Similar Errors & Avg.Length \\ \hline
        CSCD-NS & 5,000 & 2,554 & 2,497 & 57.45 \\ \hline
        \textbf{\textit{Test}} & \#Sent & \#Errors & \#Phonetically Similar Errors & Avg.Length \\ \hline
        CSCD-NS & 5,000 & 2,528 & 2,484 & 57.63 \\
        CAR & 3245 & 1,911 & 1,500 & 43.44 \\
        COT & 993 & 486 & 341 & 40.11 \\
        ENC & 3271 & 1,787 & 1,401 & 38.30 \\
        GAM & 393 & 164 & 130 & 32.81 \\
        MEC & 1942 & 1,032 & 827 & 39.18 \\
        NEW & 5887 & 3,260 & 2,698 & 25.15 \\
        NOV & 6000 & 3,415 & 2,585 & 36.24 \\ \hline
    \end{tabular}
    \caption{Statistics of the training, development and test datasets.}
    \label{tab:data_stat}
\end{table*}

\subsection{Case Study}
\begin{CJK*}{UTF8}{gbsn} 
Table~\ref{tab:case_cscd} compares the performance of C-LLM and the original LLM. 

In the first case, although the correct mapping is from \textit{"这也(as well)"} to \textit{"这一(this)"}, the model fails to understand the relationship between the incorrect characters. It splits \textit{"这也(as well)"} into two tokens and predicts characters that do not meet phonetic constraints. 

In the second case, the original LLM should map the characters \textit{"详(comprehensive)"} and \textit{"析(analyze)"} to the word \textit{"详细(detail)"}. However, it incorrectly maps \textit{"详(comprehensive)"} to \textit{"实(accurate)"}, with the predicted characters not being phonetically similar to the source ones. These errors indicate that the original LLM lacks a clear understanding of characters and words, making it unable to accurately correct misspelled words. In contrast, C-LLM can correctly correct misspelled characters within words through character-level tokenization. 

However, the third case shows that C-LLM may also make errors when correcting single incorrect characters, indicating that there is still room for improvement in our model. For some new popular words it may be necessary to combine the RAG \cite{lewis2020retrieval} method to do error correction.

% 给一个长度不一致的例子，放到第一个。第二个是拼音。第三个是CLLM的问题就行。第三个例子很好，就说对于一些新热词可能需要结合RAG的方法来做纠错。 ✅

\end{CJK*}

\begin{CJK*}{UTF8}{gbsn} 
\begin{table*}[!ht]
    \renewcommand{\arraystretch}{1}
    \centering
    \begin{tabular}{l|l}
    \hline
        Models                 & Cases in CSCD-NS test set \\ \hline
\multirow{3}{*}{Original}   & Src: 这\textcolor{red}{也}  /  更新  /  ，  /  让 ...  This \textcolor{red}{also} update allows ...\\
                       & Ref: 这 \textcolor{blue}{一}  /  更新  /  ，  /  让 ... \textcolor{blue}{This} update allows ... \\
                       & Pre: 这  /  \textcolor{red}{此}  /  更新  /  ，  /  让 ... This \textcolor{red}{this} update allows ... \\ \hdashline
\multirow{3}{*}{C-LLM} & Src: 这  /  \textcolor{red}{也}  /  更  /  新   /   ，  /  让...  This \textcolor{red}{also} update allows ...\\
                       & Ref: 这  /  \textcolor{blue}{一}  /  更新  /  ，  /  让... \textcolor{blue}{This} update allows ...\\
                       & Pre: 这  /  \textcolor{blue}{一}  /  更  /  新  /  ，  /  让 ... \textcolor{blue}{This} update allows ... \\ \hline
\multirow{3}{*}{Original}   & Src: 可  /  查询  /  详  /  \textcolor{red}{析}  /  数据  /  信息  \quad  Can query \textcolor{red}{analyzied} data information \\
                       & Ref: 可/ 查询/ 详\textcolor{blue}{细}/ 数据/ 信息 \quad   Can query \textcolor{blue}{detailed} data information \\
                       & Pre: 可  /  查询  /  详  /  \textcolor{red}{实}  /  数据  /  信息  \quad Can query \textcolor{red}{accurate} data information \\ \hdashline
\multirow{3}{*}{C-LLM} & Src: 可/  查/  询/  详/  \textcolor{red}{析}  /  数  /  据  /  信  /  息  \quad  Can query \textcolor{red}{analyzied} data information \\
                       & Ref: 可  /  查  /  询  /  详  /  \textcolor{blue}{细}  /  数  /  据  /  信  /  息 \quad  Can query \textcolor{blue}{detailed} data information \\
                       & Pre: 可  /  查  /  询  /  详  /  \textcolor{blue}{细}  /  数  /  据  /  信  /  息   \quad  Can query \textcolor{blue}{detailed} data information \\ \hline
\multirow{3}{*}{Original}   & Src: 关注  /  微信  /  \textcolor{red}{火}  /  下载  /  都有  /  机会  \quad Follow WeChat \textcolor{red}{fire} download for a chance \\
                       & Ref: 关注  /  微信  /  \textcolor{blue}{或}  /  下载  /  都有  /  机会 \quad Follow WeChat \textcolor{blue}{or} download for a chance \\
                       & Pre: 关注  /  微信  /  \textcolor{blue}{或}  /  下载  /  都有  /  机会  \quad Follow WeChat \textcolor{blue}{or} download for a chance \\ \hdashline
\multirow{3}{*}{C-LLM} & Src:  关  /注  /微  /信  /\textcolor{red}{火}  /下  /载  /都  /有  /机  /会 \quad Follow WeChat \textcolor{red}{fire} download for a chance \\
                       & Ref:  关  /注  /微  /信  /\textcolor{blue}{或}  /下  /载  /都  /有  /机  /会  \quad Follow WeChat \textcolor{blue}{or} download for a chance \\
                       & Pre: 关  /注  /微  /信  /\textcolor{red}{号}  /下  /载  /都  /有  /机  /会  
 \quad Follow WeChat \textcolor{red}{account} download for a chance \\ \hline
    \end{tabular}
    \caption{Case study of correction results between models C-LLM and Original LLM (with 14B parameters) on the CSCD-NS test set. We mark the \textcolor{red}{wrong}/\textcolor{blue}{correct} characters.}
    \label{tab:case_cscd}
\end{table*}
\end{CJK*}

\end{document}